\documentclass[letterpaper, 10 pt, conference]{ieeeconf}  

\IEEEoverridecommandlockouts                              

\usepackage{mathptmx} 
\usepackage{amsmath} 
\usepackage{amssymb}  
\usepackage{multirow}
\usepackage{graphicx}
\usepackage{hyperref}
\usepackage{color}

\title{\LARGE \bf
Force-Aware Residual DAgger via Trajectory Editing for Precision Insertion with Impedance Control
}

\author{
Yiou Huang$^{\dagger}$,
Ning Ma$^{\dagger}$,
Weichu Zhao,
Zinuo Liu,
Jun Sun,
Qiufeng Wang$^{*}$,
Yaran Chen$^{*}$%
\thanks{
All authors are with Xi'an Jiaotong-Liverpool University, Suzhou, China.
$^{\dagger}$Equal contribution.
$^{*}$Corresponding authors.
Correspondence to: \texttt{Yaran.Chen@xjtlu.edu.cn}.
This work was supported by the Research Development Fund (RDF) of Xi’an Jiaotong-Liverpool University (XJTLU) under Grant No. RDF-24-02-028 and Supported by Suzhou Science and Technology Development Planning Programme (Grant No. ZXL2025310).
}
}

\begin{document}

\maketitle
\thispagestyle{empty}
\pagestyle{empty}


\begin{abstract}
Imitation learning (IL) has shown strong potential for contact-rich precision insertion tasks. However, its practical deployment is often hindered by covariate shift and the need for continuous expert monitoring to recover from failures during execution. 
In this paper, we propose Trajectory Editing Residual Dataset Aggregation (TER-DAgger), a scalable and force-aware human-in-the-loop imitation learning framework that mitigates covariate shift by learning residual policies through optimization-based trajectory editing. This approach smoothly fuses policy rollouts with human corrective trajectories, providing consistent and stable supervision.
Second, we introduce a force-aware failure anticipation mechanism that triggers human intervention only when discrepancies arise between predicted and measured end-effector forces, significantly reducing the requirement for continuous expert monitoring.
Third, all learned policies are executed within a Cartesian impedance control framework, ensuring compliant and safe behavior during contact-rich interactions.
Extensive experiments in both simulation and real-world precision insertion tasks show that TER-DAgger improves the average success rate by over 37\% compared to behavior cloning, human-guided correction, retraining, and fine-tuning baselines, demonstrating its effectiveness in mitigating covariate shift and enabling scalable deployment in contact-rich manipulation.

\end{abstract}

\section{INTRODUCTION}
Precision insertion tasks, such as electronic component assembly and tight-tolerance part mating, are a cornerstone of modern industrial automation. These tasks are contact-rich and highly sensitive to geometric misalignment and environmental uncertainty, where small deviations can cause excessive contact forces, jamming, or damage.

Imitation Learning (IL) enables robots to acquire complex manipulation skills from human demonstrations, as demonstrated by Action Chunking with Transformers (ACT)~\cite{zhao2023learning}, Diffusion Policy (DP)~\cite{chi2025diffusion}, and Vision-Language-Action (VLA) models~\cite{kim2024openvla, black2024pi0}. However, most IL approaches rely on vision and proprioception and execute policies via position-based control, limiting compliance during physical contact.

Recent works incorporate force, torque, or tactile sensing into learning-based manipulation~\cite{yu2025forcevla, hao2025tla, zhang2025vtla, cheng2025omnivtla, huang2026tactile}, and some integrate IL policies with impedance control to improve safety~\cite{kamijo2024learning, ge2025filic}. However, they do not explicitly address distribution shift during real-world deployment.

In practice, learned policies inevitably encounter out-of-distribution states due to perception noise, modeling error, or contact uncertainty, which in contact-rich manipulation often manifest as abnormal interaction forces and can quickly lead to unsafe behavior.

Human-in-the-loop approaches, such as DAgger~\cite{ross2011reduction} and HG-DAgger~\cite{kelly2019hg}, address this issue by incorporating expert feedback during execution, but they require continuous human supervision and scale poorly. Subsequent methods~\cite{zhang2016query, hoque2021lazydagger, hoque2021thriftydagger} reduce expert burden by predicting when intervention is needed, at the cost of additional models and training complexity. Moreover, abrupt switching between autonomous execution and human control often introduces significant distribution shift. Recent work~\cite{xu2025compliant} mitigates this issue by learning residual corrections, but still relies on continuous expert monitoring.

We propose Trajectory Editing Residual Dataset Aggregation (TER-DAgger), a scalable and force-aware imitation learning framework for contact-rich precision insertion. TER-DAgger uses optimization-based trajectory editing to smoothly fuse policy rollouts with human corrective trajectories, providing consistent residual supervision and mitigating covariate shift. Human intervention is triggered only by discrepancies between predicted and measured end-effector forces, and execution is performed within a Cartesian impedance control framework for compliant and safe interaction.

Our contributions are summarized as follows:
\begin{itemize}
\item We introduce \textbf{TER-DAgger}, a force-aware human-in-the-loop imitation learning framework designed to mitigate covariate shift in contact-rich precision insertion tasks through optimization-based trajectory editing and residual policy learning.
\item We propose a \textbf{force-aware error detection mechanism} that enables scalable human supervision without auxiliary learned intervention models.
\item We integrate the proposed framework with \textbf{Cartesian impedance control} to achieve compliant and robust precision insertion.
\end{itemize}

\begin{figure*}[t]
    \centering
    \includegraphics[width=\textwidth]{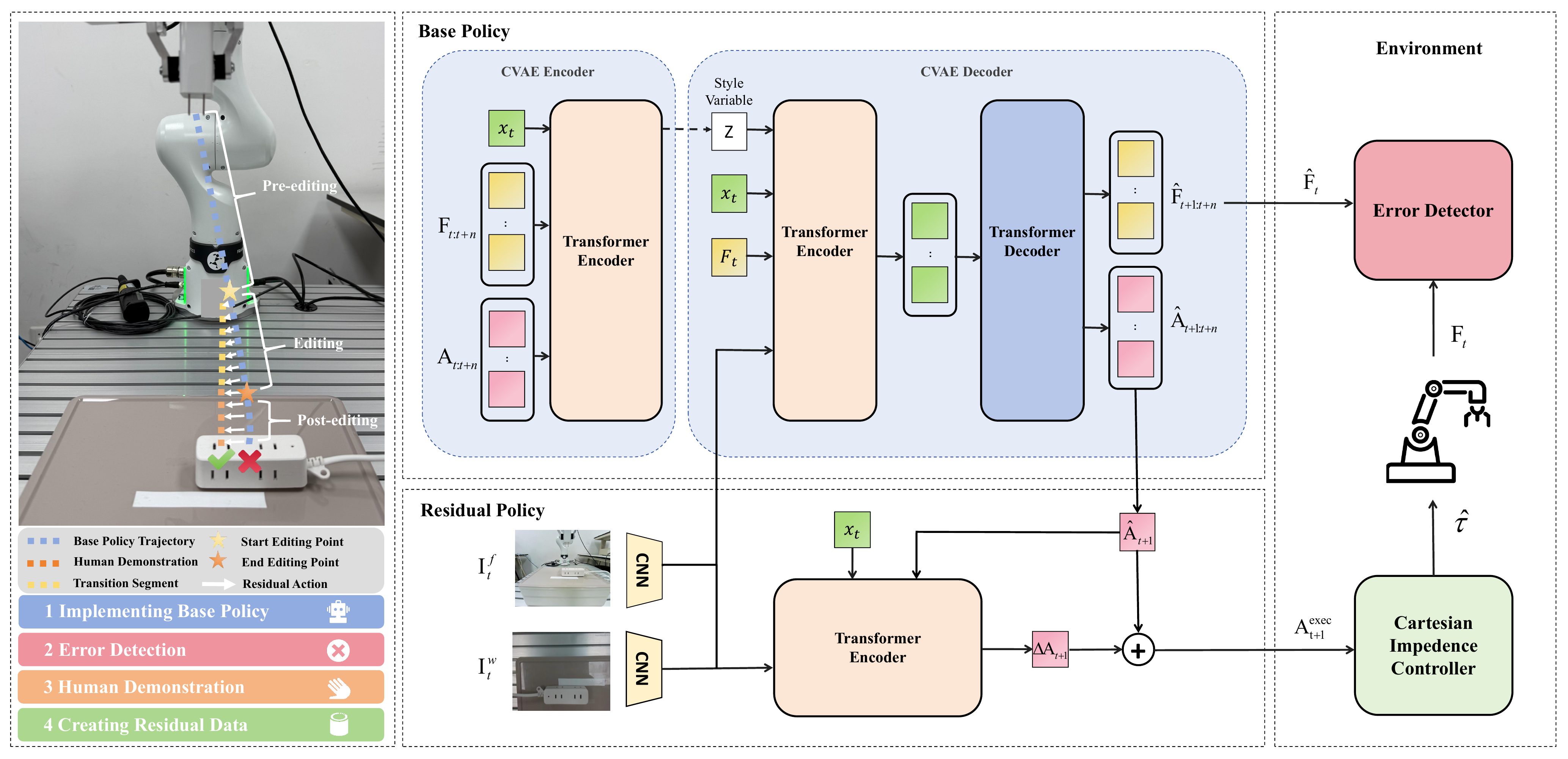}
    \caption{\textbf{(Left) TER-DAgger pipeline.}
The robot first executes the task using the base policy. When the error detector identifies a failure, execution is paused and a human provides a corrective insertion demonstration. To generate residual training data, we locate the nearest point on the base-policy trajectory to the start of the human demonstration as the editing endpoint. Together with its preceding $N-1$ points, this forms an editing segment, which is optimized to produce a smooth transition toward the human-corrected trajectory. Based on the optimized trajectory, we construct training data for the residual policy.
\textbf{(Right) Framework overview.}
The base policy (1 Hz) predicts future end-effector poses and forces from image observations, the current end-effector pose and force. The error detector (50 Hz) compares predicted and measured end-effector forces to detect failures. The residual policy (50 Hz) takes image observations, the current end-effector pose, and the next action predicted by the base policy as inputs, and predicts pose corrections. The corrected actions are executed via a Cartesian impedance controller (1 kHz).}
    \label{fig1}
\end{figure*}

\section{RELATED WORK}

\subsection{Imitation Learning for Robotic Manipulation}

IL has become a central paradigm for robotic manipulation, enabling robots to acquire complex skills from demonstrations. Early approaches such as Action Chunking with Transformers (ACT)~\cite{zhao2023learning} model manipulation as autoregressive action sequence generation. Other diffusion-based methods, e.g., Diffusion Policy (DP)~\cite{chi2025diffusion}, generate actions via conditional denoising, improving multimodal trajectory modeling. Vision-Language-Action (VLA) models~\cite{kim2024openvla, black2024pi0} further enhance generalization by leveraging large-scale vision-language pretraining.

However, most IL and VLA-based methods primarily rely on visual and proprioceptive inputs, overlooking force feedback critical for contact-rich manipulation. Recent multimodal approaches incorporate tactile and Force/Torque sensing~\cite{yu2025forcevla, hao2025tla, zhang2025vtla, cheng2025omnivtla, huang2026tactile}, yet typically execute policies with position-based control, limiting compliance and safety. Some works combine IL with impedance control~\cite{kamijo2024learning, ge2025filic}. Building on this line, we employ a Cartesian impedance controller and additionally predict future interaction forces to enable force-aware manipulation and support downstream error detection.

\subsection{Human-in-the-Loop Corrections for Imitation Learning}
The DAgger algorithm~\cite{ross2011reduction} requires experts to label actions generated by the learner’s policy without full control authority, which can pose safety risks in real-world robotic systems. HG-DAgger~\cite{kelly2019hg} improves safety by allowing direct human intervention during hazardous situations, but it requires continuous expert monitoring for each robot, limiting scalability.
To reduce expert burden, subsequent methods~\cite{zhang2016query, hoque2021lazydagger, hoque2021thriftydagger} introduce auxiliary models to predict when human takeover is necessary. While effective, these approaches increase model complexity and training overhead. Moreover, both reactive and predicted interventions often introduce abrupt state transitions, leading to significant distribution shift from the policy’s on-policy data.

Recent work explores on-policy correction strategies to address this issue. Notably, a recent study \cite{xu2025compliant} learns a residual policy from human corrective demonstrations, enabling smoother adaptation and reduced distribution shift. However, continuous human monitoring is still required, effectively restricting one expert to supervising a single robot.

In contrast, our approach also learns a residual policy but eliminates the need for continuous human attention. We automatically trigger human intervention based on the discrepancy between predicted and measured interaction forces, enabling one expert to supervise multiple robots. Additionally, an optimization-based trajectory blending scheme smoothly fuses nominal and corrective trajectories, avoiding abrupt control switching.

\subsection{Out-of-Distribution Detection}
Out-of-distribution (OOD) detection is critical to ensuring the reliability and safety of robotic systems, aiming to identify inputs that deviate from the training data distribution. Numerous approaches have been proposed to detect OOD states using deep neural network–based architectures. One line of research formulates error detection as a supervised learning problem by directly training a failure or error prediction model \cite{hendrycks2018deep}. Another widely studied direction leverages uncertainty estimation techniques—such as Bayesian neural networks \cite{gal2016dropout}, ensemble methods \cite{lakshminarayanan2017simple}, or distributional representations \cite{lee2017training}—to flag potential OOD states when predictive uncertainty is high. Alternatively, reconstruction-based methods employ reconstruction loss or Kullback--Leibler (KL) divergence of Conditional Variational AutoEncoder (CVAE) \cite{wong2022error} to identify anomalies that deviate from in-distribution patterns. In this work, we propose a straightforward yet effective indicator for OOD detection: the discrepancy between the predicted end-effector force from the base policy and the real end-effector force measured during execution.

\section{METHOD}
\label{sec:method}
The overall framework is shown in Fig.~\ref{fig1} (Right). The system consists of a force-aware base policy, an error detector, a residual policy, and a Cartesian impedance controller.

During execution, the base policy (1 Hz) predicts receding-horizon end-effector poses and interaction forces from visual observations and current states. An error detector (50 Hz) compares predicted and measured forces to detect out-of-distribution contact states and trigger human intervention. The residual policy (50 Hz) takes the same inputs along with the next base action and outputs a pose residual. The executed command is the sum of the base action and residual correction, tracked by a Cartesian impedance controller (1 kHz) for compliant interaction.

The following subsections describe each component in detail.

\subsection{Force-Aware Base Policy}
\label{sec::base_policy}

We adopt a Transformer-based architecture as the base policy. 
Following ACT~\cite{zhao2023learning}, both the encoder and decoder are implemented using CVAE Transformers.

To improve performance in contact-rich manipulation and enhance interaction awareness, we explicitly incorporate the external end-effector wrench into the policy. The external end-effector force is defined as
\begin{equation}
F_t = [f_{x,t}, f_{y,t}, f_{z,t}, \tau_{x,t}, \tau_{y,t}, \tau_{z,t}]^\top .
\end{equation}
It is estimated from measured joint torques using the pseudoinverse of the Jacobian transpose:
\begin{equation}
F_t = J(\theta_t)^{T+} \tau_t ,
\end{equation}
where $J(\theta_t)$ denotes the manipulator Jacobian at configuration $\theta_t$, $(\cdot)^{T+}$ represents the pseudoinverse of the Jacobian transpose, and $\tau_t$ is the vector of external joint torques.

\paragraph{Encoder.}
The encoder takes as input the current Cartesian end-effector pose $x_t$, the action sequence $A_{t:t+n}$, and the end-effector force sequence $F_{t:t+n}$. 
The Cartesian pose is defined as $x_t = [p_t, q_t]$, where $p_t \in \mathbb{R}^3$ denotes the position and $q_t \in \mathbb{S}^3$ is a unit quaternion representing orientation. 
The encoder outputs the mean and variance of a latent style variable $z$, which captures task-specific motion patterns and contact dynamics.

\paragraph{Decoder.}
The decoder conditions on the latent variable $z$, the current pose $x_t$, the current end-effector force $F_t$, and two RGB images $I_t^{f}, I_t^{w}$ of size $480 \times 640 \times 3$ captured from the front and wrist camera. 
Each image is encoded using a ResNet-18 backbone to extract visual features.

Unlike conventional action-only policies, the decoder jointly predicts the future action sequence $\hat{A}_{t+1:t+n}$ and the future end-effector force sequence $\hat{F}_{t+1:t+n}$. 
By explicitly modeling future interaction forces, the policy learns a force-consistent latent representation, improving stability and robustness in contact-rich scenarios.

\paragraph{Execution Strategy.}
The base policy operates at 1\,Hz. At each inference step, it predicts a horizon of $n = 100$ steps (corresponding to 2\,s), while the manipulator executes only the first $n_e = 50$ steps before replanning. 
This receding-horizon execution scheme enhances robustness against modeling errors and external disturbances while maintaining long-horizon consistency.

\subsection{Error Detection}
\label{sec:error_detection}
To detect OOD states during the insertion process, we exploit the discrepancy between the predicted and measured end-effector forces. At each timestep, the force prediction $\hat{F}_t$ produced by the base policy, which is compared against the actual measured force $F_t$.

We quantify the force prediction error using the $\ell_1$ norm:
\begin{equation}
e_t = \left\| \hat{F}_t - F_t \right\|_1.
\end{equation}

An error is triggered when the prediction error exceeds a predefined threshold $c$:
\begin{equation}
e_t > c.
\end{equation}

If this condition holds, the current state is classified as out-of-distribution. Intuitively, large force prediction errors indicate that the current contact dynamics deviate from the training distribution, which typically corresponds to unexpected collisions, misalignment, or task failure. Thus, the force prediction discrepancy serves as an implicit uncertainty estimator for reliable error detection in contact-rich manipulation.

\subsection{Trajectory Editing Residual DAgger}

To incorporate human corrections after OOD detection, we propose a trajectory editing residual DAgger framework (Fig.~\ref{fig1}, Left).

When a failure is detected, execution is paused and a short corrective demonstration is collected. Rather than re-demonstrating the full task, we align the demonstration to the nearest point on the base trajectory and locally optimize a preceding segment to form a smooth transition. The resulting corrected trajectory is then used to construct supervised residual training data.

The framework comprises three components:
(1) a residual correction policy,
(2) local corrected trajectory construction, and
(3) residual training data generation.

\subsubsection{Residual Policy}

The residual policy adopts a lightweight two-layer Transformer encoder (hidden dimension $h_r = 256$, $n_r = 8$ attention heads). At time step $t$, it takes the following inputs:
\begin{itemize}
    \item the current Cartesian end-effector pose $x_t$,
    \item the next action predicted by the base policy $\hat{A}_{t+1}$,
    \item two RGB images $I_t^{f}, I_t^{w}$ of resolution $480 \times 640 \times 3$.
\end{itemize}

Each image is encoded using a shared ResNet-18 backbone.
The residual policy outputs a corrective action
\begin{equation}
\Delta A_{t+1} = [\Delta p_{t+1}, \Delta q_{t+1}],
\end{equation}
and the executed action becomes
\begin{equation}
A^{\text{exec}}_{t+1} = \hat{A}_{t+1} + \Delta A_{t+1}.
\end{equation}


\subsubsection{Corrected Trajectory Construction}
\label{sec:correct_traj}
Let the base policy predict a trajectory
\begin{equation}
X_b = [x^b_0, x^b_1, \dots, x^b_{n_b-1}].
\end{equation}

When an OOD state is detected, we temporarily set the Cartesian impedance
stiffness to zero and collect a short corrective human demonstration
\begin{equation}
X_h = [x^h_0, x^h_1, \dots, x^h_{n_h-1}].
\end{equation}

Instead of re-demonstrating the full task, we locally edit the base trajectory
near the intervention point.

\paragraph{Nearest-Point Alignment}

We first find the closest point on $X_b$ to the initial human pose $x^h_0$:
\begin{equation}
\begin{aligned}
k^* &= \arg\min_k D(k), \\
D(k) &= \omega_p d_p(x^b_k, x^h_0)
+ \omega_q d_q(x^b_k, x^h_0).
\end{aligned}
\end{equation}

The position and orientation distances are defined as
\begin{equation}
d_p(x^b_k, x^h_0) = \| p^{b}_k - p^{h}_0 \|_2,
\end{equation}
\begin{equation}
d_q(x^b_k, x^h_0) = 1 - |\langle q^{b}_k, q^{h}_0 \rangle|.
\end{equation}

We use $\omega_p=1.0$ and $\omega_q=0.5$.

\paragraph{Local Trajectory Optimization}

We extract a local segment of length $N$:
\begin{equation}
X_b^{\text{seg}} =
\{x^b_{k^*-N}, \dots, x^b_{k^*}\}.
\end{equation}

We optimize this segment to generate a smooth transition
\begin{equation}
\tilde{X} =
\{\tilde{x}_{k^*-N}, \dots, \tilde{x}_{k^*}\},
\end{equation}
by solving
\begin{equation}
\begin{aligned}
\min_{\tilde{X}} \quad
&
\mathcal{L}_{\text{fid}}
+
\lambda_s \mathcal{L}_{\text{smooth}}
+
\lambda_e \mathcal{L}_{\text{end}} \\
\text{s.t.} \quad
&
\tilde{x}_{k^*} = x^h_0.
\end{aligned}
\end{equation}

The objective consists of three terms:

\textbf{Fidelity term}
\begin{equation}
\mathcal{L}_{\text{fid}}
=
\sum_{i=k^*-N}^{k^*}
\left(
\| p_i - p^b_i \|_2^2
+
\lambda_q^{f}
\left(
1 - |\langle q_i, q^b_i \rangle|
\right)
\right),
\end{equation}
which preserves similarity to the original base trajectory.

\textbf{Smoothness term}
\begin{equation}
\mathcal{L}_{\text{smooth}}
=
\sum_{i=k^*-N+1}^{k^*}
\left(
\| p_i - p_{i-1} \|_2^2
+
\lambda_q^{s}
\left(
1 - |\langle q_i, q_{i-1} \rangle|
\right)
\right),
\end{equation}
which enforces local motion continuity.

\textbf{Endpoint consistency term}
\begin{equation}
\mathcal{L}_{\text{end}}
=
\| p_{k^*} - p^h_0 \|_2^2
+
\lambda_q^{e}
\left(
1 - |\langle q_{k^*}, q^h_0 \rangle|
\right),
\end{equation}
which ensures alignment with the human corrective pose.

We use
\[
\lambda_s = 1.0, \quad
\lambda_e = 1000.0, \quad
\lambda_q^f = \lambda_q^s = \lambda_q^e = 0.5.
\]

The final corrected trajectory is
\begin{equation}
\label{corrected_traj}
X_{\text{correct}} =
\{
x^b_0, \dots, x^b_{k^*-N-1},
\tilde{X},
x^h_1, \dots, x^h_{n_h-1}
\}.
\end{equation}

\subsubsection{Residual Training Data Generation}
\label{sec:residual_data}
Given the corrected trajectory $X_{\text{correct}}$, we construct supervised tuples
\begin{equation}
(x_t, I_t^{f}, I_t^{w}, \hat{A}_{t+1}, \Delta A_{t+1})
\end{equation}

The dataset is divided into four regions according to the trajectory structure.

\paragraph{Pre-editing residual samples}

For
\[
t \in [0, \dots, k^*-N-1],
\quad x_t = x^b_t,
\]
the corrected trajectory coincides with the original base trajectory.
Therefore, no correction is required:
\begin{equation}
\Delta A_{t+1} = 0.
\end{equation}

This teaches the residual policy to remain inactive under nominal in-distribution states.

\paragraph{Transition residual samples}

For
\[
t \in [k^*-N, \dots, k^*-1],
\quad x_t = x^b_t,
\]
the target next pose is given by the optimized transition trajectory $\tilde{x}_{t+1}$.
The residual label is defined as
\begin{equation}
\Delta A_{t+1}
=
\tilde{x}_{t+1} - \hat{A}_{t+1}.
\end{equation}

This segment enables the residual policy to smoothly deform the base trajectory
toward the human corrective pose.

\paragraph{Human demonstration residual samples}

For
\[
t \in [0, \dots, n_h-1],
\quad x_t = x^h_t,
\]
we evaluate the base policy prediction $\hat{A}^{\text{base}}_{t+1}$ under the same
receding-horizon execution scheme.
The residual label is
\begin{equation}
\Delta A_{t+1}
=
x^h_{t+1}
-
\hat{A}^{\text{base}}_{t+1}.
\end{equation}

This teaches the residual policy to reproduce the human corrective behavior
relative to the base prediction.

\paragraph{Post-editing residual samples}

For
\[
t \in [k^*, \dots, n_b-1],
\quad x_t = x^b_t,
\]
we associate each base action $\hat{A}_{t+1}$ with the nearest pose in the human
demonstration trajectory:
\begin{equation}
\tilde{A}_{t+1}
=
\arg\min_{x^h_i}
\Big(
d_p(x^h_i, \hat{A}_{t+1})
+ 0.5 d_q(x^h_i, \hat{A}_{t+1})
\Big).
\end{equation}
The residual label becomes
\begin{equation}
\Delta A_{t+1}
=
\tilde{A}_{t+1} - \hat{A}_{t+1}.
\end{equation}

This encourages consistency with the corrective intent.

\subsection{Cartesian Impedance Controller}

To ensure compliant and safe manipulation during physical contact, we adopt a Cartesian impedance controller as the low-level controller to track the predicted trajectory. Cartesian impedance control regulates the dynamic interaction between motion and external forces by shaping the apparent mechanical behavior of the end-effector as a virtual mass–spring–damper system in Cartesian space.

Specifically, the desired interaction behavior is defined as
\begin{equation}
F_{\text{imp}}
=
K (x_d - x)
+
D (\dot{x}_d - \dot{x}),
\end{equation}
where $F_{\text{imp}} = [f_x, f_y, f_z, \tau_x, \tau_y, \tau_z]^\top$ denotes the desired Cartesian interaction wrench generated by the impedance controller, $x_d$ and $\dot{x}_d$ denote the desired Cartesian pose and velocity of the end-effector, and $x$ and $\dot{x}$ represent the measured pose and velocity, respectively. $K$ and $D$ are positive-definite stiffness and damping matrices, respectively.

To achieve real-time control at 1\,kHz with improved numerical robustness, we employ a simplified torque-level implementation that avoids explicit operational-space inertia matrix computation. The commanded joint torque is given by
\begin{equation}
\hat{\tau}
=
J(\theta)^{T}
\left(
K (x_{d} - x)
+
D (\dot{x}_{d} - \dot{x})
\right)
+
g(\theta),
\end{equation}
where $J(\theta)$ is the manipulator Jacobian matrix, $J(\theta)^{T}$ maps Cartesian wrenches to joint torques, and $g(\theta)$ denotes the gravity compensation term.

This impedance formulation enables the manipulator to remain compliant under unexpected contacts while accurately tracking the high-level trajectory predicted by the policy.

\begin{figure}[!t]
    \vspace{2mm}
    \centering
    \includegraphics[width=8.5cm]{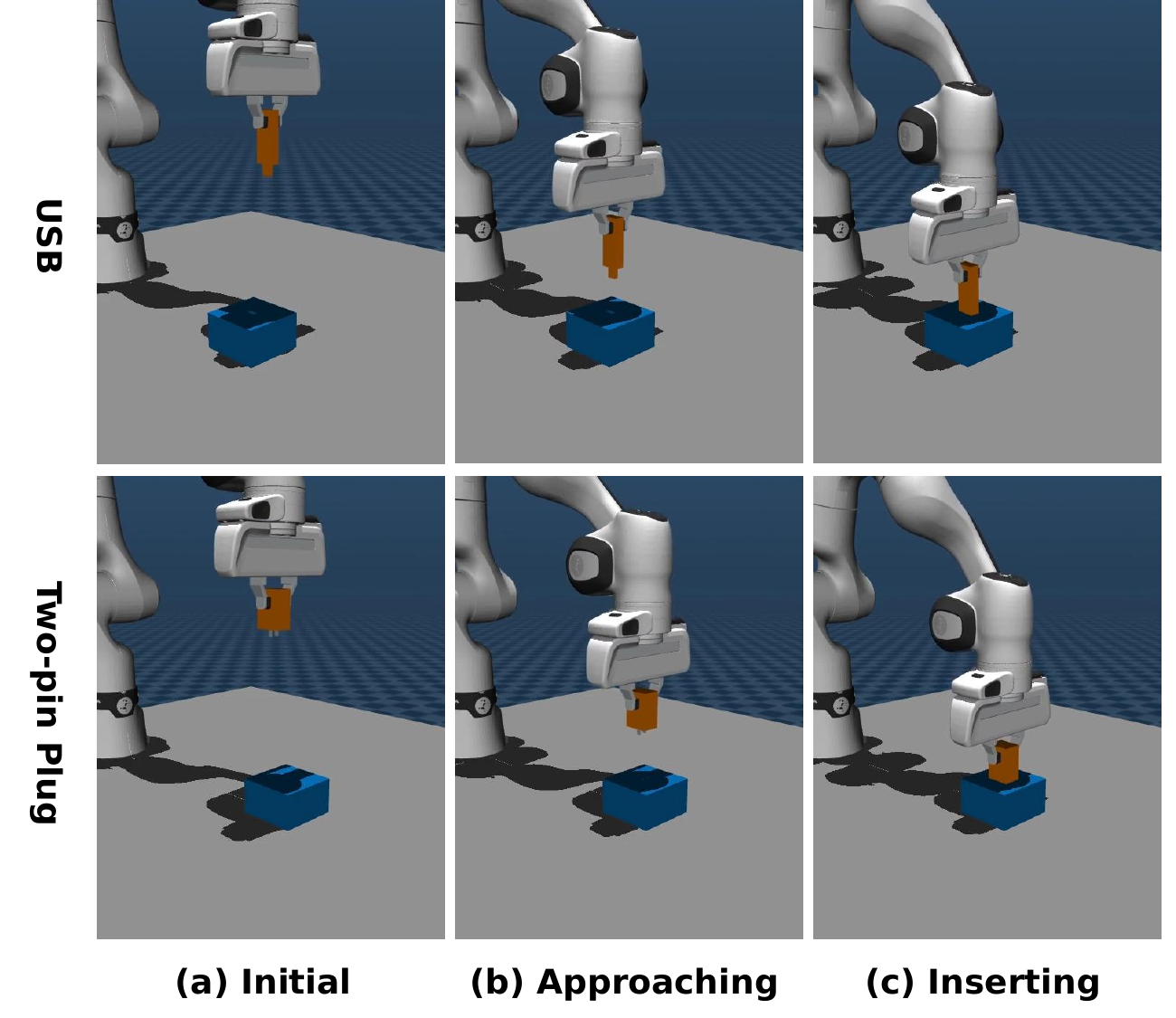}
    \caption{Simulation scene setup and insertion task process.}
    \label{fig2}
\end{figure}

\begin{figure}[!t]
    \centering
    \includegraphics[width=8.5cm]{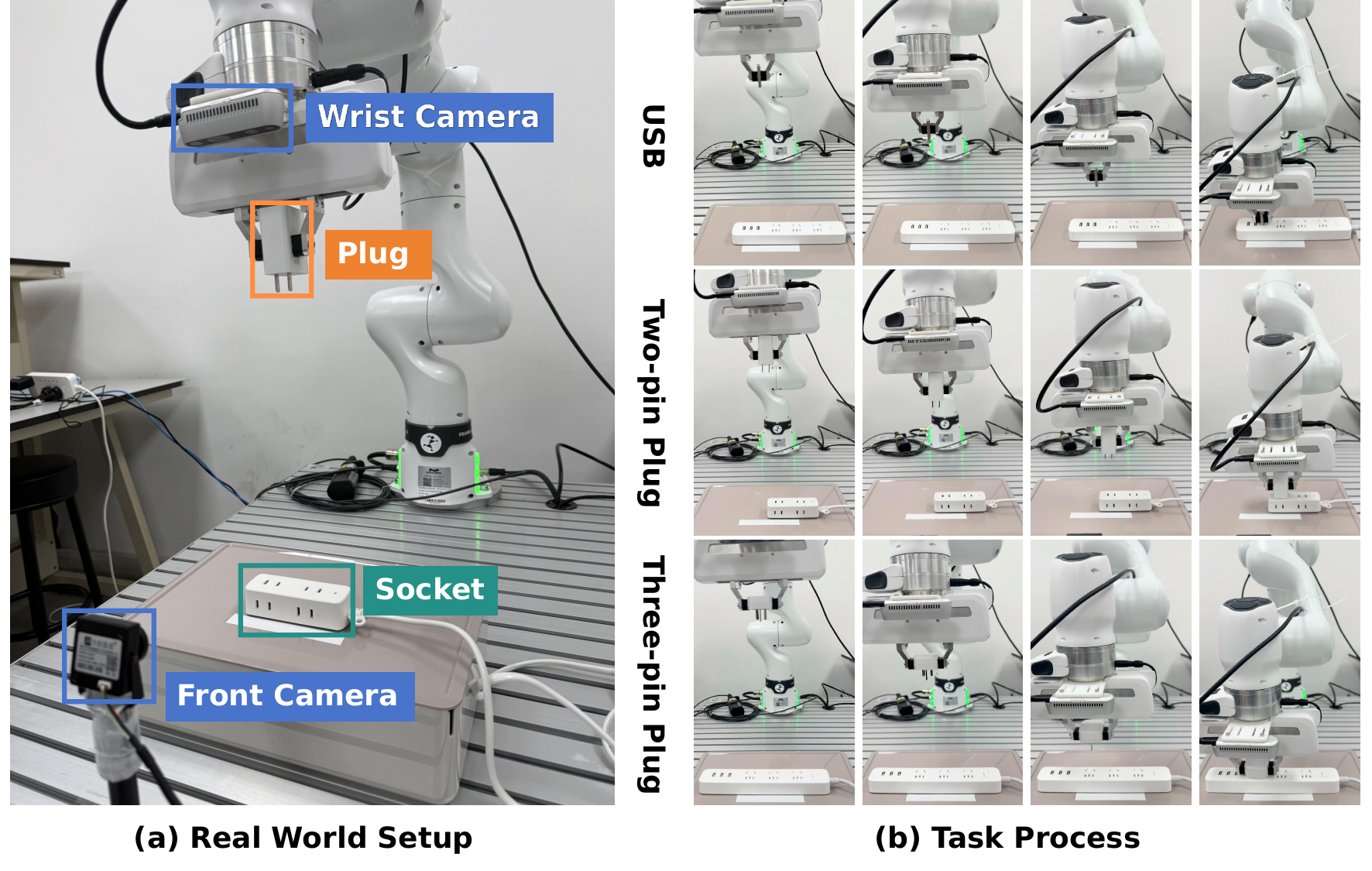}
    \caption{Real scene setup and insertion task process.}
    \label{fig3}
\end{figure}

\section{EXPERIMENTS}
Our approach is validated through experiments in simulation and on real robotic systems. We begin by outlining the experimental setups, data-collection procedures, and then present a comparative analysis.

\subsection{Simulation Setup}
Experiments are conducted in MuJoCo with a Franka Research 3 manipulator equipped with wrist and front-view RGB cameras (Fig.~\ref{fig2}).

\textbf{Tasks.}
We evaluate two insertion tasks: (1) \emph{USB insertion} and (2) \emph{two-pin plug insertion}. In both tasks, the plug is mounted on the end-effector, initialized 30\,cm above the socket. The socket position is uniformly sampled along the $y$-axis within $[-5\,\mathrm{cm}, 5\,\mathrm{cm}]$. The insertion tolerance is 0.1\,mm.

\textbf{Data Collection.}
For base policy learning, training trajectories are generated using a rule-based method. From a fixed initial pose with randomized socket positions, an approach pose 8\,cm above the target is defined, and linear interpolation generates the full trajectory. For each task, 100 trajectories are collected, including synchronized RGB images, Cartesian poses, joint torques, and end-effector forces.

For residual policy learning, 50 corrective demonstrations per task are collected. Each trajectory interpolates from 8\,cm above the correct target to the final pose and includes synchronized RGB images and Cartesian poses.

\subsection{Real-World Setup}
Real-world experiments use the same Franka robot with a wrist-mounted Intel RealSense D415 and a fixed front-view RGB camera (Fig.~\ref{fig3}).

\textbf{Tasks.}
We evaluate USB, two-pin plug, and three-pin plug insertion. The plug is initialized 30\,cm above the socket, whose position is randomly varied along a white tape to introduce spatial diversity.

\textbf{Data Collection.}
For base policy learning, training trajectories are recorded in teach mode with zero Cartesian impedance stiffness, then replayed to log joint torques and end-effector forces. For each task, 100 trajectories are collected with synchronized RGB images, Cartesian poses, joint torques, and end-effector forces.

For residual policy learning, 50 corrective demonstrations per task are recorded. After failure, the end-effector is moved approximately 8\,cm above the correct target, from which a corrective trajectory is demonstrated, containing synchronized RGB images and Cartesian poses.

\subsection{Experiment Results}

\subsubsection{Baseline Methods for TER-DAgger}

To rigorously evaluate the effectiveness of the proposed TER-DAgger framework, we compare it against several representative baselines in two simulated environments and three real-world precision insertion tasks. The evaluated methods are summarized as follows:

\begin{itemize}
    \item \textbf{ACT~\cite{zhao2023learning}:}
    The original Action Chunk Transformer serves as the foundational policy architecture of our method. It is trained using the 100 base trajectories for 2000 epochs.

    \item \textbf{FILIC~\cite{ge2025filic}:}
    A multimodal imitation learning framework that incorporates visual observations, end-effector poses, and force measurements. Given its architectural similarity to our base policy, FILIC provides a strong multimodal baseline. It is trained using the same 100 base trajectories for 2000 epochs.

    \item \textbf{HG-DAgger~\cite{kelly2019hg}:}
    A human-guided DAgger variant in which an expert monitors policy rollouts and intervenes when failure is anticipated. In our implementation, Cartesian impedance stiffness is set to zero during intervention to allow compliant correction. We collect 50 corrective demonstrations and finetune the pre-trained base policy for 200 epochs.

    \item \textbf{Retrain:}
    The base policy is retrained from scratch using a combined dataset consisting of the original 100 base trajectories and 50 corrective demonstrations (recorded following Eq.~\ref{corrected_traj}). Training is performed for 2000 epochs.

    \item \textbf{Finetune:}
    The pre-trained base policy is finetuned using only the 50 corrective demonstrations (recorded following Eq.~\ref{corrected_traj}) for 200 epochs.

    \item \textbf{TER-DAgger (Ours):}
    Our proposed two-stage framework first trains a base policy using 100 trajectories for 2000 epochs. Subsequently, 50 corrective demonstrations are collected to train a residual policy for 200 epochs, as detailed in Sec.~\ref{sec:method}.
\end{itemize}

For each task, all methods are evaluated over 50 independent trials under identical testing conditions.

\subsubsection{Experiment Results for TER-DAgger}
As shown in Table~\ref{tab1}, \textbf{TER-DAgger consistently outperforms all baseline methods} across both simulation and real-world precision insertion tasks, achieving an average success rate of \textbf{77.2\%}, which exceeds the strongest baseline (Finetune, 40.0\%) by over \textbf{37\%}.

Standard imitation learning methods exhibit severe performance degradation during deployment. The base ACT~\cite{zhao2023learning} policy performs poorly in both simulated and real environments, particularly in tasks with tight contact tolerances, confirming that covariate shift leads to rapid failure accumulation in contact-rich manipulation. Although FILIC~\cite{ge2025filic} incorporates force feedback, its limited improvement indicates that augmenting observations alone is insufficient to resolve distribution mismatch.

Human-guided approaches such as HG-DAgger~\cite{kelly2019hg} improve performance in simulation but generalize poorly to real-world tasks, suggesting that direct human takeover introduces additional distribution discontinuities that misalign correction data with the policy’s execution distribution.

Retrain and Finetune benefit from additional correction data but show inconsistent gains across tasks. Their inferior performance relative to TER-DAgger indicates that naively aggregating or fine-tuning on correction trajectories does not adequately address the state-action distribution mismatch encountered during execution.

In contrast, TER-DAgger achieves near-perfect performance in simulation (90\% and 96\%) and substantial gains in real-world experiments, reaching 96\% success in two-pin plug insertion and maintaining strong performance on the more challenging three-pin plug task (82\%). These results demonstrate that \textbf{TER-DAgger effectively mitigates covariate shift by aligning training data with on-policy execution}, resulting in robust and scalable performance for contact-rich precision insertion.

\begin{table*}[t] 
\vspace{2mm}
\small 
\centering
\caption{Success rates (\%) of different methods} 
\label{tab1} 
\begin{tabular}{c|cc|ccc|c}
\hline
\multirow{2}{*}{\textbf{Method}} & \multicolumn{2}{c|}{\textbf{Simulation}} & \multicolumn{3}{c|}{\textbf{Real}} & \multirow{2}{*}{\textbf{Average}} \\
                                 & \textbf{USB} & \textbf{Two-pin Plug} & \textbf{USB} & \textbf{Two-pin Plug} & \textbf{Three-pin Plug} & \\
\hline
ACT~\cite{zhao2023learning} & 26\% & 22\% & 2\% & 64\% & 24\% & 27.6\% \\
FILIC~\cite{ge2025filic} & 20\% & 30\% & 10\% & 44\% & 24\% & 25.6\% \\
HG-DAgger~\cite{kelly2019hg} & 50\% & 44\% & 16\% & 36\% & 12\% & 31.6\% \\
Retrain & 48\% & 56\% & 0\% & 36\% & 22\% & 32.4\% \\
Finetune & 68\% & 54\% & 10\% & 28\% & 40\% & 40.0\% \\
\textbf{TER-DAgger (Ours)} & \textbf{90\%} & \textbf{96\%} & \textbf{22\%} & \textbf{96\%} & \textbf{82\%} & \textbf{77.2\%} \\
\hline
\end{tabular}
\end{table*}

\subsubsection{Baseline Methods for Error Detection}
To evaluate the effectiveness of the proposed force-aware failure detection mechanism, we compare it against several baseline error detection strategies derived from the base policy architecture and commonly used uncertainty or prediction-based metrics.

\begin{itemize}
    \item \textbf{KL Loss:} The KL divergence between the output of the CVAE encoder $z$ of the base policy and the standard normal distribution $\mathcal{N}(0, I)$ is used as an uncertainty-based error indicator.
    
    \item \textbf{Reconstruction Loss:} The reconstruction error of the base policy is employed as a baseline metric, reflecting the discrepancy between predicted and reconstructed actions under the learned latent representation.
    
    \item \textbf{Position Prediction Error:} This baseline measures the Euclidean error between the Cartesian action pose predicted by the base policy and the current end-effector pose, capturing deviations in the predicted motion command.
    
    \item \textbf{Force Prediction Error (Ours):} The details are described in \ref{sec:error_detection}
\end{itemize}

For all methods, task-specific thresholds are selected as reported in Table~\ref{tab3}. The threshold for each metric is chosen to guarantee 100\% recall of failure cases while maximizing precision, ensuring fair comparison under the same safety constraint. For each task, 50 test samples are used for evaluation.

\subsubsection{Experiment Results for Error Detection}
As shown in Table~\ref{tab2}, the proposed force prediction error consistently achieves the \textbf{highest precision across all tasks while maintaining 100\% recall}. In contrast, KL loss, reconstruction loss, and position prediction error exhibit substantially lower precision despite achieving full recall, leading to a higher rate of false positives.

This gap is particularly pronounced in real-world tasks, where uncertainty-based metrics such as KL loss and reconstruction error suffer from significant precision degradation. Position prediction error also performs inconsistently, as pose deviations do not reliably correlate with contact failures in contact-rich insertion scenarios.

In contrast, force prediction error remains highly discriminative across both simulation and real-world settings, achieving an average precision of \textbf{98.8\%} while preserving complete recall. These results demonstrate that the proposed method can \textbf{reliably detect all failure cases with minimal false alarms}, enabling human intervention only when necessary.

By maximizing precision under the constraint of full recall, the proposed force-aware error detection mechanism ensures the lowest possible human monitoring cost while maintaining safety during contact-rich manipulation.

\begin{table*}[t] 
\small 
\centering
\caption{Comparison of error detection methods} 
\label{tab2} 
\begin{tabular}{cccccc}
\hline
\textbf{Task}                                                    & \textbf{Metric}                                            & \textbf{KL Loss}                                     & \textbf{\begin{tabular}[c]{@{}c@{}}Reconstruction \\ Loss\end{tabular}} & \textbf{\begin{tabular}[c]{@{}c@{}}Position Prediction \\ Error\end{tabular}} & \textbf{\begin{tabular}[c]{@{}c@{}}Force Prediction \\ Error (Ours)\end{tabular}} \\ \hline
Two-pin Plug                                                     & \begin{tabular}[c]{@{}c@{}}Precision $\uparrow$\\ Recall $\uparrow$\end{tabular} & \begin{tabular}[c]{@{}c@{}}90\\ 100\end{tabular}     & \begin{tabular}[c]{@{}c@{}}98\\ 100\end{tabular}                        & \begin{tabular}[c]{@{}c@{}}76\\ 100\end{tabular}                              & \textbf{\begin{tabular}[c]{@{}c@{}}100\\ 100\end{tabular}}                        \\ \hline
USB                                                              & \begin{tabular}[c]{@{}c@{}}Precision $\uparrow$\\ Recall $\uparrow$\end{tabular} & \begin{tabular}[c]{@{}c@{}}74\\ 100\end{tabular}     & \begin{tabular}[c]{@{}c@{}}78\\ 100\end{tabular}                        & \begin{tabular}[c]{@{}c@{}}78\\ 100\end{tabular}                              & \textbf{\begin{tabular}[c]{@{}c@{}}100\\ 100\end{tabular}}                        \\ \hline
\begin{tabular}[c]{@{}c@{}}Two-pin Plug \\ (Real)\end{tabular}   & \begin{tabular}[c]{@{}c@{}}Precision $\uparrow$\\ Recall $\uparrow$\end{tabular} & \begin{tabular}[c]{@{}c@{}}50\\ 100\end{tabular}     & \begin{tabular}[c]{@{}c@{}}40\\ 100\end{tabular}                        & \begin{tabular}[c]{@{}c@{}}36\\ 100\end{tabular}                              & \textbf{\begin{tabular}[c]{@{}c@{}}100\\ 100\end{tabular}}                        \\ \hline
\begin{tabular}[c]{@{}c@{}}USB \\ (Real)\end{tabular}            & \begin{tabular}[c]{@{}c@{}}Precision $\uparrow$\\ Recall $\uparrow$\end{tabular} & \begin{tabular}[c]{@{}c@{}}100\\ 100\end{tabular}    & \begin{tabular}[c]{@{}c@{}}96\\ 100\end{tabular}                        & \begin{tabular}[c]{@{}c@{}}100\\ 100\end{tabular}                             & \textbf{\begin{tabular}[c]{@{}c@{}}100\\ 100\end{tabular}}                        \\ \hline
\begin{tabular}[c]{@{}c@{}}Three-pin Plug \\ (Real)\end{tabular} & \begin{tabular}[c]{@{}c@{}}Precision $\uparrow$\\ Recall $\uparrow$\end{tabular} & \begin{tabular}[c]{@{}c@{}}64\\ 100\end{tabular}     & \begin{tabular}[c]{@{}c@{}}64\\ 100\end{tabular}                        & \begin{tabular}[c]{@{}c@{}}76\\ 100\end{tabular}                              & \textbf{\begin{tabular}[c]{@{}c@{}}94\\ 100\end{tabular}}                         \\ \hline
Average                                                          & \begin{tabular}[c]{@{}c@{}}Precision $\uparrow$\\ Recall $\uparrow$\end{tabular} & \begin{tabular}[c]{@{}c@{}}75.6\\ 100.0\end{tabular} & \begin{tabular}[c]{@{}c@{}}75.2\\ 100.0\end{tabular}                    & \begin{tabular}[c]{@{}c@{}}73.2\\ 100.0\end{tabular}                          & \textbf{\begin{tabular}[c]{@{}c@{}}98.8\\ 100.0\end{tabular}}                     \\ \hline
\end{tabular}
\end{table*}

\begin{table}[t]
\small
\centering
\caption{Thresholds of Error Detection Methods}
\label{tab3}
\resizebox{\columnwidth}{!}{
\begin{tabular}{ccccc}
\hline
\textbf{Task} & \textbf{KL} & \textbf{Recon.} & \textbf{Pos.} & \textbf{Force (Ours)} \\
              & \textbf{Loss} & \textbf{Loss} & \textbf{Err.} & \textbf{Err.} \\
\hline
USB                     & 0.00056 & 0.00162 & 0.012 & 11.0 \\
Two-pin Plug            & 0.0011  & 0.00086 & 0.012 & 13.0 \\
USB (Real)              & 0.014   & 0.00078 & 0.018 & 16.0 \\
Two-pin Plug (Real)     & 0.00109 & 0.005   & 0.024 & 15.0 \\
Three-pin Plug (Real)   & 0.0015  & 0.0006  & 0.025 & 14.0 \\
\hline
\end{tabular}
}
\end{table}

\subsection{Ablation Studies}

We conduct ablation studies on the two-pin plug insertion task in simulation to analyze three key design choices of the proposed framework.

\paragraph{Adding End-effector Force as Input for Base Policy}
We first evaluate the effect of incorporating end-effector force into the base policy, as described in Section~\ref{sec::base_policy}. 
As shown in Table~\ref{tab4}, augmenting the ACT~\cite{zhao2023learning} architecture with force input consistently improves performance across all tasks, increasing the average success rate from 27.6\% to 32.4\%.
This improvement indicates that explicit force information provides valuable interaction cues that are not fully observable from vision and pose alone, enabling the policy to better reason about contact states in precision insertion.

\paragraph{Training Samples for Residual Policy}
We next analyze the contribution of different components of the residual training data described in Section~\ref{sec:residual_data}. 
The pre-editing residual samples are always included to enforce zero residual under nominal in-distribution states.
We selectively ablate the remaining three components—transition samples, human demonstration samples, and post-editing samples—and report the results in Table~\ref{tab5}.

Using only individual components yields limited improvement over the base policy, with post-editing samples providing the most significant single contribution.
Combining different components leads to progressively better performance, and the best result is achieved when all components are used jointly, reaching a success rate of 96\%.
These results demonstrate that the proposed residual data construction strategy is complementary: transition samples enable smooth correction onset, demonstration samples capture corrective intent, and post-editing samples enforce long-horizon consistency.

\paragraph{Number of Points for Optimization}
Finally, we study the effect of the number of optimization points $N$ used in local trajectory editing, as introduced in Section~\ref{sec:correct_traj}. 
As shown in Table~\ref{tab6}, performance is robust within a moderate range of $N$, peaking at $N=20$.
Using too few points limits the smoothness of the transition, while overly long segments degrade performance by over-constraining the trajectory.
This result suggests that local trajectory editing benefits from a balanced optimization horizon that is sufficiently long to ensure smoothness without sacrificing flexibility.

\begin{table*}[t] 
\vspace{2mm}
\small 
\centering
\caption{Ablation studies of adding end-effector force} 
\label{tab4} 
\begin{tabular}{c|cc|ccc|c}
\hline
\multirow{2}{*}{\textbf{Method}} & \multicolumn{2}{c|}{\textbf{Simulation}} & \multicolumn{3}{c|}{\textbf{Real}} & \multirow{2}{*}{\textbf{Average}} \\
                                 & \textbf{USB} & \textbf{Two-pin Plug} & \textbf{USB} & \textbf{Two-pin Plug} & \textbf{Three-pin Plug} & \\
\hline
ACT~\cite{zhao2023learning} & 26\% & 22\% & 2\% & 64\% & 24\% & 27.6\% \\
\textbf{Base Policy (Ours)} & \textbf{34\%} & \textbf{28\%} & \textbf{4\%} & \textbf{68\%} & \textbf{28\%} & \textbf{32.4\%} \\
\hline
\end{tabular}
\end{table*}




\begin{table}[]
\centering
\caption{Ablation studies of three parts of residual samples} 
\label{tab5} 
\begin{tabular}{lc}
\hline
\textbf{Residual Training Samples} & \textbf{Success Rate} \\ \hline
Base Policy (Ours)                  & 28\%                  \\
Transition                & 32\%                  \\
Demonstration            & 34\%                  \\
Post-editing        & 66\%                  \\
Transition + Demonstration         & 42\%                  \\
Transition + Post-editing    & 56\%                  \\
Demonstration + Post-editing & 92\%                  \\
\textbf{TER-DAgger (Ours)}                         & \textbf{96}\%         \\ \hline
\end{tabular}
\end{table}

\begin{table}[]
\centering
\caption{Ablation studies of number of points for optimization} 
\label{tab6} 
\begin{tabular}{cc}
\hline
\textbf{Number of Points} & \textbf{Success Rate} \\ \hline
10                        & 94\%                  \\
20                        & \textbf{96\%}         \\
30                        & 88\%                  \\
40                        & 80\%                  \\ \hline
\end{tabular}
\end{table}

\section{CONCLUSIONS}

We presented TER-DAgger, a force-aware human-in-the-loop imitation learning framework designed to mitigate covariate shift in contact-rich precision insertion tasks. By aligning supervision with on-policy execution, TER-DAgger prevents the performance degradation typical of standard imitation learning in real-world deployment.

The framework detects out-of-distribution contact states via force prediction discrepancies and triggers human intervention only when necessary, reducing monitoring cost. Local trajectory editing incorporates corrective demonstrations as residual supervision, ensuring smooth integration without distribution discontinuities.

Experiments in both simulation and real-world settings show that TER-DAgger consistently outperforms behavior cloning, human-guided correction, retraining, and fine-tuning baselines. Ablation studies confirm the critical role of force modeling, residual data construction, and local trajectory optimization in reducing compounding errors.

Future work will extend TER-DAgger to more complex precision assembly tasks and explore its scalability in broader robotic manipulation scenarios.

\addtolength{\textheight}{-12cm}   




\bibliographystyle{IEEEtran}
\bibliography{IEEEabrv,ref}

\end{document}